\documentclass[11pt]{article}
\usepackage{acl-hlt2011}
\usepackage{subfigure,url}
\usepackage[mathscr]{euscript}
\usepackage[OT1,T1]{fontenc}
\newif{\ifcitewithnumbers}
\citewithnumbersfalse

\ifcitewithnumbers
    
     \usepackage[numbers,square,sort&compress]{natbib}
    \renewcommand{\newcite}[2][]{\citet[#1]{#2}}
\else

     \providecommand{\citet}[1]{\newcite{#1}}
  \providecommand{\citep}[2][]{\cite[#1]{#2}}
\fi

\newif{\ifsplitsumfig}
\splitsumfigfalse

\ifsplitsumfig
\newcommand{\sumfigsplitter}{\\}
\newcommand{\sumwidth}{2.5in}
\else
\newcommand{\sumfigsplitter}{}
\newcommand{\sumwidth}{1.5in}
\fi

\newif{\ifackfirstinitial}
\ackfirstinitialfalse

\usepackage{times}
\usepackage{latexsym}
\usepackage{xspace}
\usepackage{graphicx}
\usepackage{pdfsync}

\usepackage[dvipsnames]{color}
\usepackage[overload]{textcase}
\usepackage{amssymb}
\usepackage{centernot}
\usepackage{enumitem}

\newcommand{\dataurl}{\url{http://www.cs.cornell.edu/~cristian/movies}}

\newcommand{\emfeat}[1]{\fbox{#1}}
\newcommand{\Narrativeimportance}{Narrative importance\xspace}
\newcommand{\narrativeimportance}{narrative importance\xspace}
\newcommand{\turn}{exchange\xspace}
\newcommand{\turns}{exchanges\xspace}

\newcommand{\fight}{2.8in}

\newcommand{\scriptwriters}{scriptwriters\xspace}
\newcommand{\scriptwriter}{scriptwriter\xspace}
\newcommand{\accommodation}{convergence\xspace} 
\newcommand{\Accommodation}{Convergence\xspace}
\newcommand{\converge}{converge\xspace}

\newcommand{\Male}{\textbf{M}ale\xspace}
\newcommand{\Female}{\textbf{F}emale\xspace}
\newcommand{\Lead}{1st resp.} 
\newcommand{\Second}{2nd resp.}

\newcommand{\article}{Article}
\newcommand{\articles}{Articles}

\newcommand{\dimension}{trigger family\xspace}

\newcommand{\dimensions}{trigger families\xspace}

\newcommand{\accfun}
{Conv}

\newcommand{\journal}{J.\xspace}
\newcommand{\jof}{\journal \xspace}

\newcommand{\Aperson}{A}
\newcommand{\Bperson}{B}

\newcommand{\trig}{t}
\newcommand{\indt}[1]{#1^\trig}
\newcommand{\trigger}{trigger\xspace}
\newcommand{\Aon}{\indt{\Autt} = 1}
\newcommand{\Bon}{\indt{\Brep} = 1}
\newcommand{\indOff}[1]{\indt{#1} = 0}
\newcommand{\Boff}{\indOff{\Brep}}
\newcommand{\Aoff}{\indOff{\Autt}}
\newcommand{\Autt}{a}
\newcommand{\Brep}{b_{\hookrightarrow\Autt}}

\title{Chameleons in imagined conversations:  A new approach to  understanding coordination of linguistic style in  dialogs
}

\author
{Cristian Danescu-Niculescu-Mizil and Lillian Lee \\ Department of
Computer Science,  Cornell University \\ {cristian@cs.cornell.edu, llee@cs.cornell.edu}}

\begin{document}

\maketitle
\begin{abstract}
Conversational participants tend to immediately and unconsciously adapt 
to each other's language styles:  
a
speaker will even adjust the number 
of articles and other function words in their next 
utterance
in response 
to the number in their partner's immediately preceding utterance.
This striking level of coordination is 
thought
to have 
arisen
as a way to achieve social goals, such as
gaining approval or emphasizing difference in status.  But has the adaptation mechanism
become so deeply embedded in the language-generation process as to
become a reflex?  We argue that fictional dialogs offer a way to study
this question, since
authors create the conversations but don't
receive the social
 benefits (rather, the imagined characters do).  
Indeed, we find significant
coordination
across
many families of function words in
our large movie-script corpus.
We also report suggestive preliminary findings on the effects of  gender and other features;
e.g., surprisingly, for articles, 
on average,
characters adapt more to females than to males.

\end{abstract}

\section{Introduction}\label{sec:intro}

{
{\em ``...it is dangerous to base any sociolinguistic argumentation on the
evidence of language in fictional texts only''} (Bleichenbacher
(2008), crediting Mare\v{s}
(2000))\nocite{Bleichenbacher:2008p3633} \nocite{Mares:2000a}
}

\bigskip

{\em The chameleon effect} 
is the
``nonconscious mimicry of the postures, mannerisms, facial expressions, and other behaviors of one's interaction partners
'' \cite{Chartrand1999893}.\footnote{The term
 is  a reference to the movie {\em Zelig}, wherein a
 ``human chameleon'' 
uncontrollably takes on the characteristics of those around him. 
The term is meant to contrast with ``aping'', a word 
connoting
intentional imitation.

Related terms 
include
adaptation, 
alignment, entrainment, 
priming, and Du Bois' dialogic syntax.} 
For example, if one conversational participant crosses their arms, 
their partner
 often unconsciously crosses their arms
as well. 
The effect
occurs for 
language, too,
ranging from matching of acoustic features such as 
accent, speech rate, and pitch
\cite{Giles:1991,Chartrand2009219}
to lexico-syntactic priming across 
adjacent or nearby utterances
\cite{Bock:1986a,Pickering:2004p5042,Ward:2007p5044,Reitter+Keller+Moore:2011a}.

Our work focuses on
adjacent-utterance
coordination with respect to
classes of 
 function words.
To exemplify the phenomenon, we discuss two short conversations.

\medskip

 \noindent $\bullet$ {\it First example:} The following exchange from 
the movie ``The Getaway'' \shortcite{getaway} 
demonstrates 
quantifier
coordination.

\smallskip

\hspace*{-.15in}{\small
\begin{tabular}{ll}
Doc:~ \ At \emfeat{least} you were outside.\\ 
Carol: It doesn't make \emfeat{much} difference where you are [...] 
\end{tabular}
}

\smallskip

\noindent Note that ``Carol'' used a quantifier, 
one that is different
than the one ``Doc'' employed.  Also, notice that ``Carol''
could just as well have replied in a way that
doesn't include a quantifier, for example, ``It doesn't really matter
where you are...''.

\medskip

 \noindent $\bullet$ {\it Second example:}  \newcite{Levelt:1982p3608} report
 an experiment involving preposition coordination.  Shopkeepers who
 were called and asked ``\emfeat{At} what time does your shop
 close?''  were
significantly
  more likely to 
say
``\emfeat{At} five o'clock'' 
than ``five o'clock''.\footnote{%
This is an example of lexical matching manifested as part
of syntactic coordination.}

Coordination of function-word class has been previously documented in
several settings
\cite{Niederhoffer:2002p2556,Taylor:2008p3340,Ireland01122010,Gonzales:2010p3341},
the largest-scale study being on Twitter  \cite{Danescu-Niculescu-Mizil+al:11a}.

\medskip

\paragraph{%
Problem setting
} 
People don't consciously track 
function
words
\cite{Levelt:1982p3608,Segalowitz+Lane:2004,vanPetten+Kutas:1991} ---
it's not 
easy to answer the question, ``how many
prepositions
were there in the sentence I just said?''.
Therefore, it is quite striking that
humans
nonetheless instantly adapt to
each other's function-word rates.
Indeed, there is 
active debate
regarding what mechanisms cause nonconscious
coordination \cite{Ireland01122010,Branigan:2010p3605}.   

One 
line of thought is that \accommodation represents a 
social strategy\footnote{
In fact,
social signaling may also be
   the evolutionary cause of chameleons' color-changing ability
  (Stuart-Fox et al., 2008).
 \nocite{Stuart:2008}
}
whose aim is to gain the other's social approval
\cite{Giles:2008,Street:1982} or enhance  the other's comprehension
\cite{Clark:1996,Bortfeld:1997}.\footnote{For the purpose of our
  discussion, we are conflating 
social-approval and audience-design hypotheses under the
category of 
\textit{social strategy}.}
This hypothesis is supported by studies showing that coordination is 
affected by a number of social factors, including relative social
status \cite{Natale:1975a,Gregory19961231,Thakerar+al:1982} and gender role
\cite{Bilous:1988,Namy:2002p3609,Ireland:2010}.

But an important question is whether the adaptation mechanism has  become
so deeply embedded in the language-generation process
as to 
have transformed into
a reflex not requiring any social triggering.\footnote{
This hypothesis relates to characterizations of alignment 
 as an unmediated mechanism \cite{Pickering:2004p5042}.}
{
Indeed, it has been argued that unconscious mimicry is partly innate
\cite{Chartrand1999893}, perhaps due to evolutionary pressure to
foster relationships \cite{Lakin:10.1023/A:1025389814290}.}  

To answer this question,
we take a radical approach: 
we consider a setting in which
the persons {\em generating} the
coordinating
dialog
are different from those {\em engaged} in the dialog
(and standing to reap the social benefits) ---
imagined conversations, specifically, 
scripted
movie dialogs.

\paragraph{Life is beautiful, but cinema is paradise}
A priori, it is not clear
that movie conversations would exhibit \accommodation.
Dialogs between movie characters 
are not truthful representations of real-life conversations.  They often are ``too carefully polished, too rhythmically balanced, too self-consciously artful'' \cite{Kozloff}, 
due to practical
and artistic
 constraints and scriptwriting practice
\cite{McKee}. 
\nocite{xkcd}
For example, mundane phenomena such as stuttering and word repetitions
are
generally
nonexistent
on the big screen.
Moreover, writers have many goals to accomplish, including the
need to advance the plot, reveal character, make jokes as funny as
possible, and so on, all incurring a cognitive load. 

So, the question arises: do scripted movie dialogs, in spite of this
quasi-artificiality and the aforementioned generation/engagement gap, exhibit the real-life phenomenon of stylistic \accommodation?  
When imagining dialogs, do \scriptwriters (nonconsciously\footnote{The
  phenomenon of real-life language convergence is not widely known
  among screenplay authors (Beth F. Milles, professor of acting and
  directing, personal communication).}) adjust the respondent's replies to echo the initiator's use of articles, prepositions, and other apparently minor
aspects of lexical choice?   
According to our results,
this is indeed the case, 
which
 has fascinating implications.

First,
this provides evidence that
coordination,
assumed to be driven by social motivations,
has become
so deeply embedded into our ideas of what conversations ``sound like''
that 
the phenomenon
occurs
even when the person generating the
dialog is not the recipient of the social benefits.\footnote{Although some writers may perhaps imagine themselves "in the shoes" of the
recipients, recall that authors generally don't include in their
scripts the 
repetitions and ungrammaticalities of "real-life" speech.}

Second,
{movies can be seen as a controlled environment in which 
preconceptions about the relation between communication  patterns and the social features of the participants 
can be studied.}  
This gives us the 
opportunity to understand how
people 
(\scriptwriters)
nonconsciously {\em expect} 
\accommodation 
to relate to factors such as
gender, status and relation
 type.
Are female characters 
thought to accommodate more to male characters than vice-versa?

Furthermore, movie scripts constitute a corpus that is especially convenient because
meta-features like gender
can be more or less readily obtained.

\paragraph{Contributions}  We check for 
\accommodation
in a
corpus of 
roughly
250,000
conversational \turns 
from
movie scripts
(available at \dataurl).
Specifically, 
we examine
the set of nine families of stylistic features previously utilized by \newcite{Ireland01122010},
and find 
a statistically significant 
\accommodation effect for all these families.
We
thereby
provide
evidence 
that language coordination is so 
implanted within our conception of
conversational behavior that, even if such coordination is socially
motivated, it is exhibited even when the person generating the
language in question is
not receiving any of the presumed social advantages.

We also 
study 
the effects of
gender,
\narrativeimportance, 
and hostility.
Intriguingly, we find that these factors indeed ``\mbox{affect}'' movie characters'
linguistic behavior;
since the characters aren't real, and control of stylistic lexical
choice is largely nonconscious, the effects of 
these factors
can only be springing from 
 patterns existing in the
\scriptwriters' minds.

Our findings, by enhancing our understanding of linguistic adaptation effects in stylistic word choice and its relation to various socially relevant factors,  may in the future aid in practical applications. 
Such an understanding
would give us insight into how and what kinds of
language 
coordination
yield more satisfying interactions
--- convergence has been already shown to 
enhance
communication in organizational contexts \cite{Bourhis:1991},
psychotherapy \cite{Ferrara:1991}, care
of the  mentally disabled \cite{Hamilton:1991}, and police-community
interactions \cite{Giles:2006p2557}.
Moreover,  a deeper understanding can aid human-computer interaction
by informing the construction of natural-language generation systems, 
since
people are often more satisfied with encounters exhibiting
appropriate linguistic convergence \cite{Bradac:1988p3600,Baaren:2003p3606},
even when the other conversational participant is known to be a
computer 
\cite{Nass:2000:CSM:332040.332452,Branigan:2010p3605}.

\section{Related work not already mentioned}
\label{sec:relwork}

\paragraph{Linguistic style and human characteristics}

Using 
stylistic 
(i.e., non-topical)
elements
like articles 
and prepositions
to
characterize the utterer in some way
 has a long history, including in
authorship attribution \cite{Mosteller+Wallace:84a,Juola:2008},
personality-type
classification
\cite{Argamon:2005p3097,Oberlander+Gill:2006,mairesse:using},
gender 
categorization \cite{Koppel:2002p3082,Mukherjee:2010p2452,Herring:2006p2209},
identification of interactional style
\cite{jurafsky-ranganath-mcfarland:2009:NAACLHLT09,ranganath-jurafsky-mcfarland:2009:EMNLP},
and recognizing deceptive language 
\cite{Hancock:5192831,mihalcea-strapparava:2009:Short}.

\paragraph{Imagined conversations}
There has been work in the NLP community applying computational
techniques
to fiction, scripts, and other types of text containing
imagined conversations. 
For example,
one recent project identifies
conversational networks in 
novels,
with the goal of evaluating various literary theories
\cite{elson-dames-mckeown:2010:ACL,Elson+McKeown:2010}.  
Movie
scripts were used as word-sense-disambiguation evaluation data as part
of an effort to generate computer animation from the scripts
\cite{Ye:2006:ALTA2006}.
\newcite{Sonderegger2010} employed a corpus of English poetry to study
the relationship between pronunciation and network structure.
\newcite{Rayson+Wilson+Leech:2002} computed part-of-speech frequencies
for imaginative writing in the British National Corpus, finding a
typology gradient progressing from conversation to imaginative writing
(e.g., novels)
to task-oriented speech to informative writing.  The data analyzed by
\newcite{Oberlander+Gill:2006} consisted of  emails that participants
were instructed to write by imagining that they were going to update a
good friend on their current goings-on.

\section{Movie dialogs corpus}
\label{sec:data}
To address 
the questions raised in the introduction,
we created
a large set of imagined 
conversations, starting from 
movie scripts
crawled from various 
sites.\footnote{The source of these scripts and more detail about the
  corpus are given in the README associated with the 
Cornell movie-dialogs corpus,
available at \dataurl 
~.
}
Metadata for conversation analysis and 
duplicate-script
detection involved
mostly-automatic matching of movie scripts with the IMDB movie
database; clean-up resulted in  617 unique titles tagged with genre,
release year, cast lists, and IMDB 
information. 
We then
extracted 220,579 conversational \turns
between pairs of characters engaging in at least 5 \turns,
and automatically matched these
characters to
IMDB
 to retrieve  gender
(as indicated by the designations ``actor'' or ``actress'')
 and/or 
billing-position
information when possible ($\approx$9000 characters, $\approx$3000 gender-identified and $\approx$3000 
billing-positioned).  The latter feature serves as a proxy for 
\narrativeimportance: the higher up in the 
credits, the more important the character tends to be in the film.

To the best of our knowledge, this is the largest dataset of 
(metadata-rich) imaginary conversations to date.

\section{Measuring linguistic style}
\label{sec:style}

For consistency with prior work, 
we employed the nine LIWC-derived
categories \cite{liwc} deemed by \citet{Ireland01122010} to be
processed by humans in a
generally non-conscious fashion.
The nine categories are:
articles, 
auxiliary verbs, 
conjunctions, 
high-frequency adverbs, 
impersonal pronouns, 
negations, 
personal pronouns,
prepositions, and
quantifiers 
(451 lexemes total).

It is important to note that language coordination is multimodal: it does not necessarily
occur 
simultaneously
for all features \cite{Ferrara:1991}, and
speakers may converge on some features but diverge on others \cite{Thakerar+al:1982}; for
example,  females have been found to converge on pause frequency with
male conversational partners but diverge on laughter
\cite{Bilous:1988}.

\section{Measuring \accommodation}
\label{sec:measure}

\newcite{Niederhoffer:2002p2556} use 
the correlation coefficient 
to
measure accommodation with respect to linguistic style features.
While correlation at first seems reasonable, it has some problematic
aspects
in our setting 
(we discuss these problems later)
that motivate us to employ an alternative measure.

We instead use a 
convergence
measure  introduced
in \citet{Danescu-Niculescu-Mizil+al:11a} 
that quantifies 
how much
a
given feature family $\trig$ serves as 
an {\em immediate trigger} or stimulus, meaning that one person's utterance
exhibiting 
such a  feature triggers the appearance of that feature
in the respondent's immediate reply.

For example, we might be studying whether one person $\Aperson$'s
inclusion of articles in an utterance triggers the usage of articles
in respondent $\Bperson$'s reply.  Note that this differs from asking
whether $\Bperson$ uses articles more often when talking to $\Aperson$
than when talking to other people (it is not so surprising that people
speak differently to different audiences).  This also differs from asking
whether $\Bperson$ eventually starts matching $\Aperson$'s behavior in
later utterances  within the same conversation.  We specifically want
to know whether each utterance by $\Aperson$ triggers an {\em
  immediate} change in $\Bperson$'s behavior, as such instantaneous
adaptation 
is what we consider  the most striking aspect of \accommodation,
although immediate and 
long-term
coordination are clearly related.

 We now
 describe
the  statistic we employ
to measure the extent to which person $\Bperson$ accommodates to $\Aperson$.
Consider
an arbitrary conversational 
\turn 
started by $\Aperson$, and let
$\Autt$ denote $\Aperson$'s initiating utterance and $\Brep$
denote $\Bperson$'s  
reply
to
$\Autt$.\footnote{
We use ``initiating'' and ``reply'' loosely: in our terminology,  the
conversation $\langle \Aperson$: ``Hi.'' $\Bperson$: ``Eaten?''  $\Aperson$:
``Nope.''$\rangle$ has two \turns, one initiated by $\Aperson$'s ``Hi'',
the other by $\Bperson$'s ``Eaten?''. 
} 
 Note that we use lowercase to emphasize when  we are talking about
individual utterances rather than all the utterances of the particular
person, and that thus,  the arrow in $\Brep$ indicates that we 
mean the  reply to the  specific single utterance $\Autt$.
Let $\indt{\Autt}$ be the
indicator
variable for $\Autt$ exhibiting 
$\trig$, and similarly
for $\indt{\Brep}$. 
Then,
we define the \accommodation 
$\accfun_{\Aperson,\Bperson}(\trig) $
of $\Bperson$ to $\Aperson$ 
as:

\vspace*{-.1in}
{\small
\begin{equation}
P(\Bon |\Aon) -
P(\Bon).
\label{eq:accdef}
\end{equation}\label{eq:accpair}
}

\vspace*{-.2in}
\noindent 
Note that this quantity can be negative 
(indicating \textit{divergence}).
The overall degree
 $\accfun(\trig)$ 
 to which $\trig$ serves as a \trigger is
then defined as
the expectation of $\accfun_{\Aperson,\Bperson}(\trig)$ over all 
initiator-respondent 
pairs:
\begin{equation}
\accfun(\trig) \stackrel{def}{=} E_{{\rm pairs }(\Aperson,\Bperson)} (\accfun_{\Aperson,\Bperson}(\trig)).
\end{equation}\label{eq:acc}

\paragraph{Comparison with correlation: the importance of
  asymmetry\footnote{Other asymmetric measures based on conditional
    probability of occurrence have been
    proposed for adaptation within monologues \cite{Church:2000a} and
    between conversations \cite{Stenchikova+Stent:2007a}.  Since our 
    focus
  is different, we control for different factors.}}
Why do we employ  $\accfun_{\Aperson,\Bperson}$,   Equation
(\ref{eq:accdef}), 
instead of the well-known 
correlation coefficient?
One reason is that correlation fails to
capture
an important asymmetry.
The
case where $\Aon$ but $\Boff$ represents a true failure to accommodate;
but the case where $\Aoff$ but $\Bon$
should not, at least not to the
same degree.
For example, $\Autt$ may be very short (e.g., ``What?'')
and thus not
contain an article, but we don't assume that this completely disallows
$\Bperson$ from using articles in their reply.
In other words, we are interested in whether the presence of $\trig$
acts as a \trigger,
not in
whether $\Brep$ exhibits $\trig$ if and only if $\Autt$ does, the
latter being what correlation 
detects.\footnote{
One could also speculate that it is easier for $\Bperson$ to
(unconsciously) pick up on the presence of $\trig$ than on its absence.}

It bears mentioning
that since $\indt{\Autt}$ and $\indt{\Brep}$
are binary,
a simple calculation shows that the covariance\footnote{The covariance
  of two random variables is their correlation times the product of
  their standard deviations.}
$cov(\indt{\Autt},\indt{\Brep}) = \accfun_{\Aperson,\Bperson}(\trig)
\cdot P(\Aon)$.  But, 
the two terms on the right
hand side
are not independent:  raising $ P(\Aon)$ could
cause 
$\accfun_{\Aperson,\Bperson}(\trig)$ to decrease by affecting the
first term in its definition, $P(\Bon |\Aon)$ (see 
eq.
\ref{eq:accdef}).

\section{Experimental results}
\subsection{\Accommodation exists in fictional dialogs}

\begin{figure}[t!]
\centering
\includegraphics[height=3.3in]{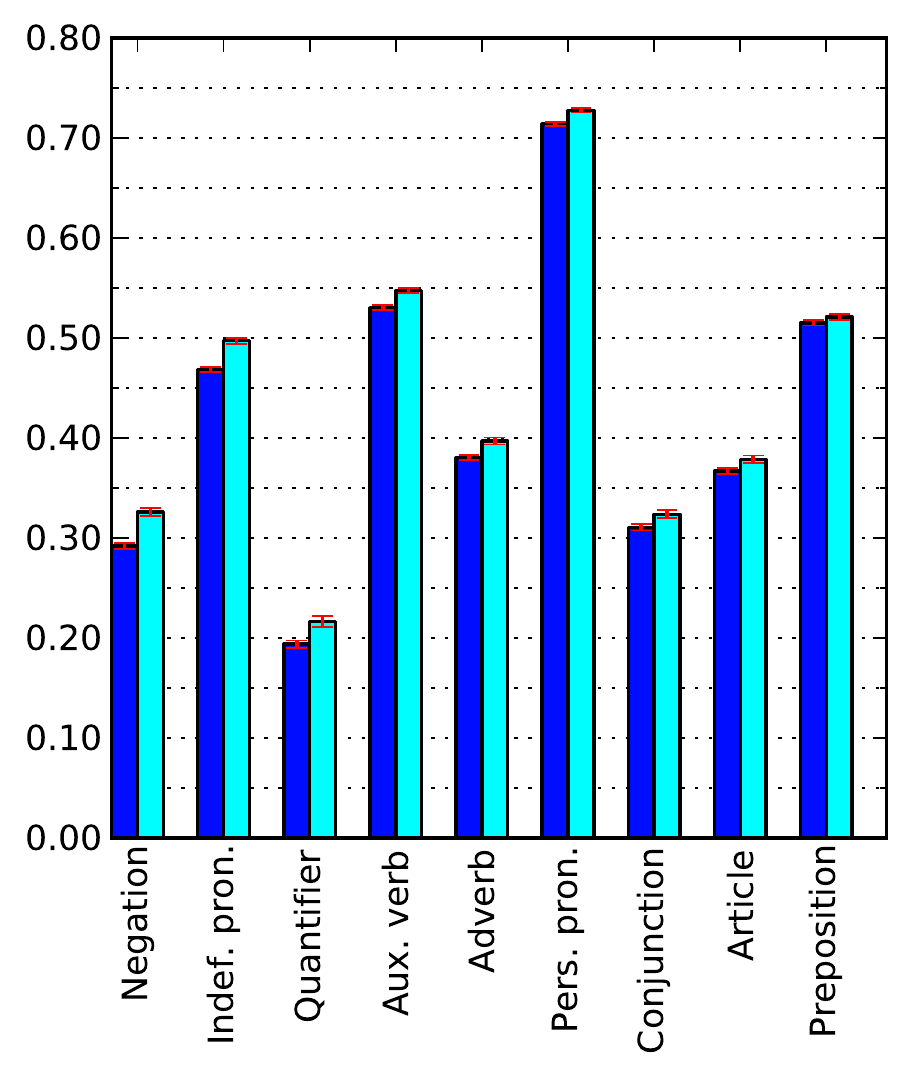}
\caption{
Implicit depiction of
\accommodation 
for each trigger family $\trig$, illustrated as the difference between
the 
means of \mbox{${P}(\Bon|\Aon)$} (right/light-blue bars) and 
\mbox{${P}(\Bon)$} (left/dark-blue bars).  
(This implicit representation allows one to see the magnitude of the
two components making up our definition of \accommodation.)
The trigger families are 
ordered by decreasing \accommodation.
All differences are statistically significant
(paired t-test). \\
In all figures in this paper, error bars represent standard error, estimated via
bootstrap resampling \cite{Koehn:2004p3868}. 
(Here, the error
bars, in red,  are  very tight.)
} 
\label{fig:acc}
\end{figure}

\begin{figure}[th!]
\includegraphics[height=\fight]{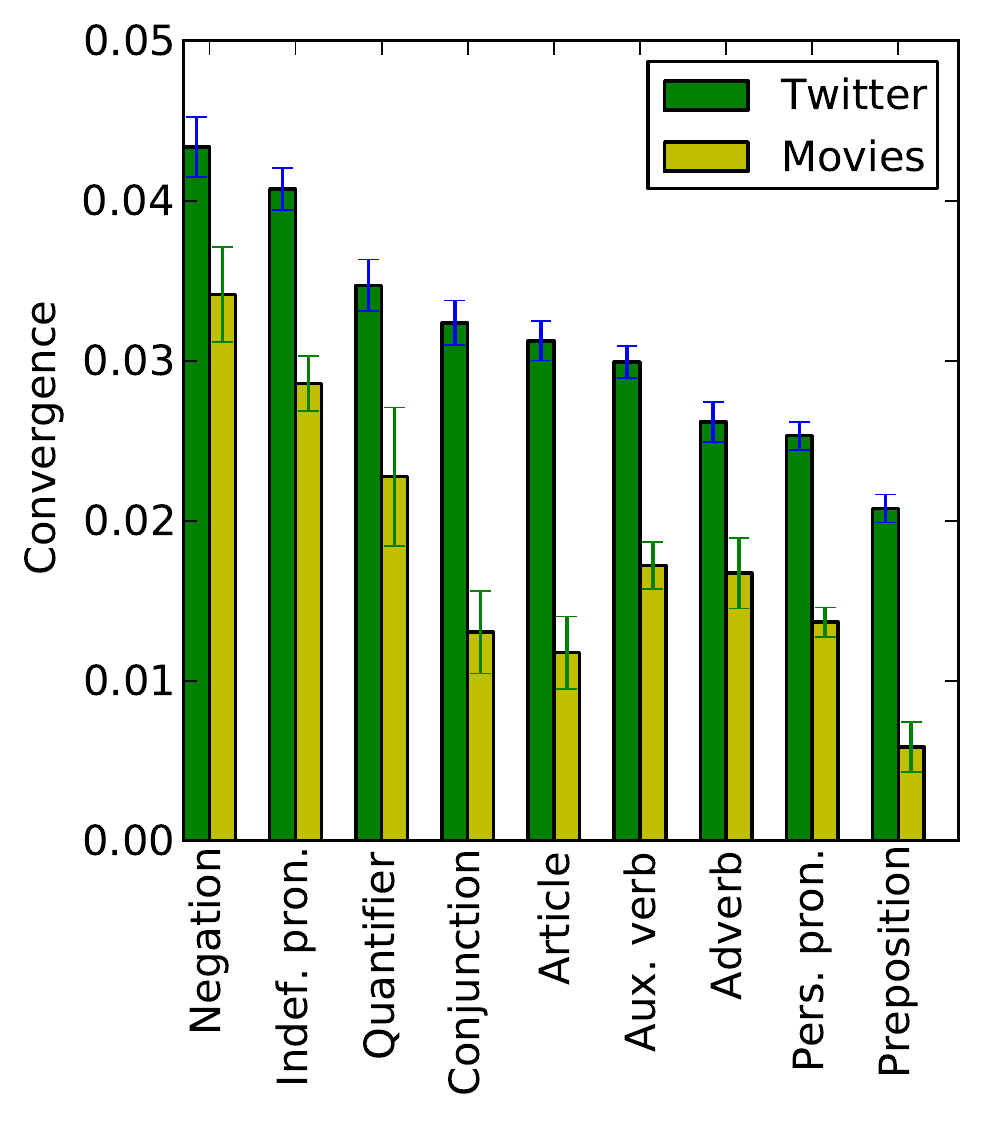}
\caption{Convergence in Twitter conversations (left bars) vs. convergence in movie dialogs (right bars;
corresponds to the difference between the two respective bars in Fig.~\ref{fig:acc}) for each trigger family.  The trigger families are ordered by decreasing convergence in Twitter.}
\label{fig:twittermovies}
\end{figure}

\begin{figure}[th!]
\includegraphics[height=\fight]{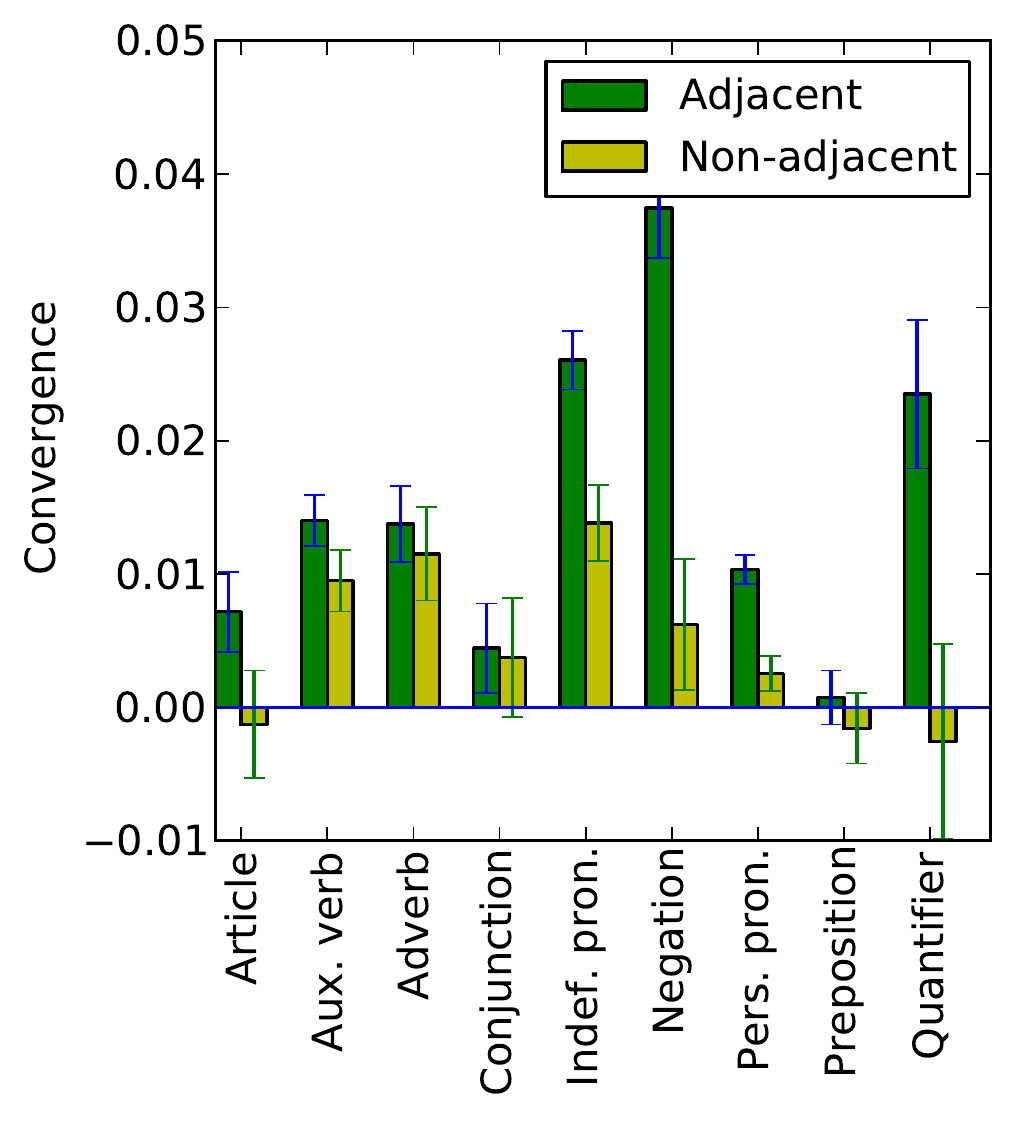}
\caption{Immediate vs. within-conversation effects 
(for conversations with at 
least
5 utterances). Suppose that we
  have a conversation $\Autt_1 \,  b_2 \, \Autt_3 \,  b_4 \,  \Autt_5 \ldots$. The
  lefthand/dark-green bars show the usual \accommodation measure,
  which involves the utterance pair $\Autt_1$ and $b_2$.  The
  righthand/mustard-green bars show \accommodation based on 
pairs like
  $\Autt_1$ and $b_4$ --- utterances in the same conversation, but not
adjacent.  We see that there is a much stronger triggering effect for
immediately adjacent utterances. }
\label{fig:decay}
\end{figure}

For each ordered pair of 
characters $(\Aperson,\Bperson)$ and for each feature family $\trig$, we
estimate 
equation
(\ref{eq:accdef})
in a straightforward manner:
the fraction of $B$'s replies to $\trig$-manifesting $\Aperson$
utterances that themselves exhibit $\trig$, minus
the fraction of all replies of $\Bperson$ to 
$\Aperson$ that 
exhibit
$\trig$.\footnote{
For each $\trig$, we discarded pairs of characters  where some
relevant count is $<10$,
 e.g., where $B$ had fewer than 10 replies
manifesting the trigger.
}
Fig. \ref{fig:acc} 
compares the average values of these two fractions (as a way of putting \accommodation values into context), showing 
positive differences
for all of the considered families of
 features
(statistically significant,
paired t-test $p$ < 0.001);
 this demonstrates that movie characters do indeed \converge to each other's linguistic style on all
considered \trigger families.\footnote{
We obtained the same qualitative results when measuring \accommodation
via the correlation coefficient, doing so for the sake of
comparability with prior work \cite{Niederhoffer:2002p2556,Taylor:2008p3340}.
}

\paragraph{Movies vs. Twitter} 
One can ask how our results on movie dialogs correspond to those for
real-life conversations. To study this, we utilize the results of
\newcite{Danescu-Niculescu-Mizil+al:11a} on a large-scale collection
of Twitter exchanges as  data on real conversational exchanges. Figure
\ref{fig:twittermovies} depicts the comparison, revealing two
interesting effects.  First, Twitter users coordinate more than movie
characters on all the trigger families we considered, which does show
that the \accommodation effect is stronger in actual interchanges.
On the other hand, from the perspective of  potentially using
imagined dialogs as proxies for real ones, it is intriguing to see that there is
generally a correspondence between how much convergence occurs in real
dialogs for a given feature family and how much convergence occurs for
that feature in imagined dialogs, although conjunctions
and articles show a bit less convergence in fictional exchanges than
this pattern would suggest.

\subsection{Potential alternative explanations}

\paragraph{Immediate vs. within-conversation effects}
An additional natural
 question is, how much are these accommodation
effects due to an immediate triggering effect, as opposed to simply
being a by-product of utterances occurring within the same
conversation?  For instance, could the results be due just to  the
topic of the conversation?

To answer this question requires measuring ``\accommodation'' between
utterances that are not adjacent, but are still in the same conversation.  To this end, we 
first restricted attention to those conversations in which there were
at least five utterances, so that they would have the structure
$\Autt_1 \, b_2 \, \Autt_3 \,  b_4 \,  \Autt_5 ...$.  We then 
measure \accommodation not between adjacent utterances,
like $\Autt_1$ and $b_2$, but where we skip an utterance, such as
the pair $\Autt_1, b_4$ or $b_2, \Autt_5$.
This helps control for
topic effects, since 
$b_4$ and $\Autt_1$ are still close and thus fairly
likely to be on the same subject.\footnote{It is true that they might be on
different topics, but in fact even $b_2$ might be on a different
subject from $\Autt_1$.}

Figure \ref{fig:decay} shows that the level of \accommodation always
falls off  after the skipped utterance, sometimes dramatically so,
thus demonstrating that the level of immediate adaptation effects we see
cannot be solely explained by the topic of conversation or other
conversation-level effects.  
These results accord with the findings of
 \newcite{Levelt:1982p3608}, where interposing ``interfering''
 questions lowered the chance of a question's preposition being echoed
 by the respondent,  and \newcite{Reitter:2006}, where the effects of
 structural priming
were shown to
decay quickly with the distance between the priming trigger and the priming target.

Towards the same end, we also performed randomization experiments in
which we shuffled the order of each participant's utterances 
in each conversation,
while
maintaining alternation between speakers.
We again observed
drop-offs in this randomized condition in comparison to immediate
\accommodation, the main focus of this paper.
\paragraph{Self-coordination} 
Could our results be explained entirely by  the 
author 
converging
to their own self, given that 
self-alignment
 has been documented
\cite{Pickering:2004p5042,Reitter:2006}?  If that were the
case, 
then the 
{\em characters} that the author is writing about should \converge to
themselves 
no more than 
they \converge to different characters.
But we ran
experiments showing that this is not the case, thus invalidating this
alternative hypothesis.  In fact, characters \converge to themselves
much more than they converge to other characters.

\begin{figure*}[th!]
\centering
\subfigure[]{
\includegraphics[height=\fight]{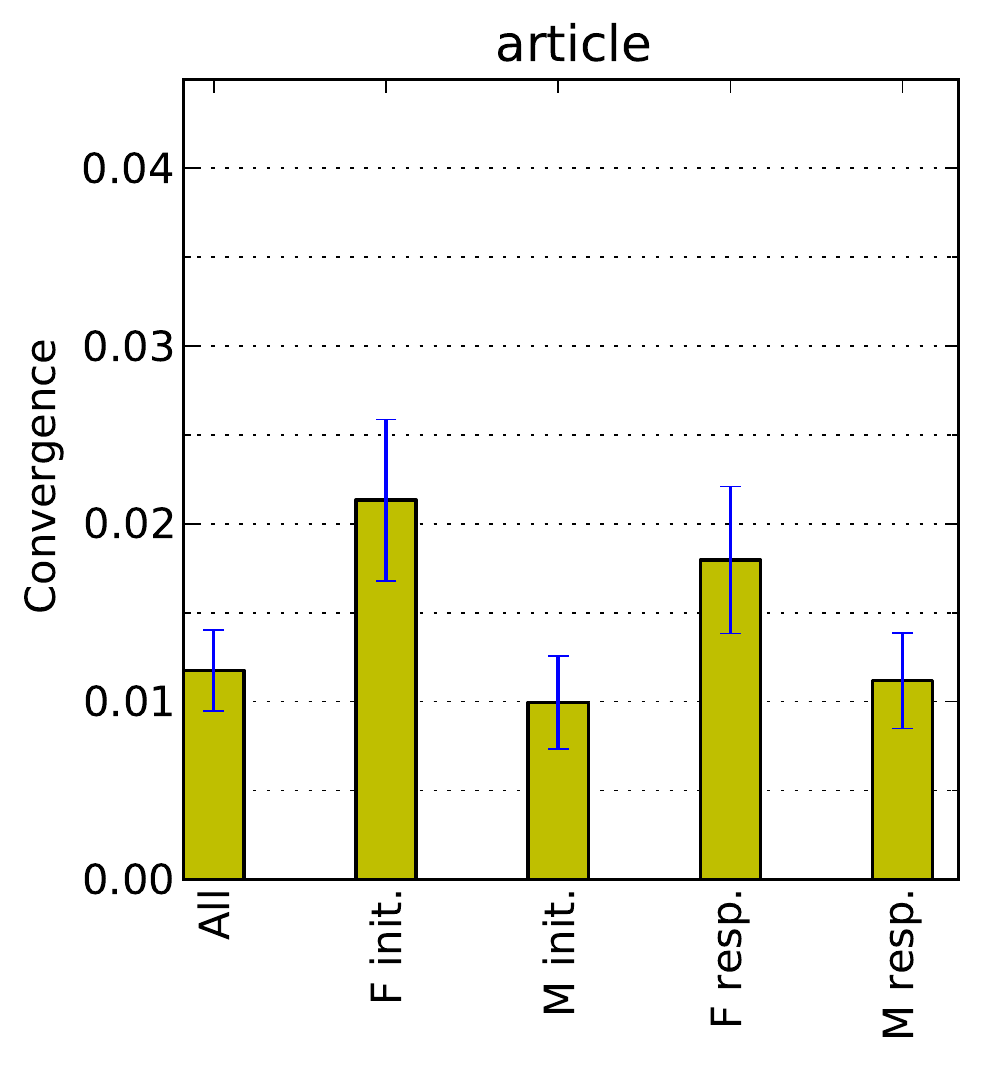}
\label{fig:gender1}
}
\subfigure[]{
\includegraphics[height=\fight]{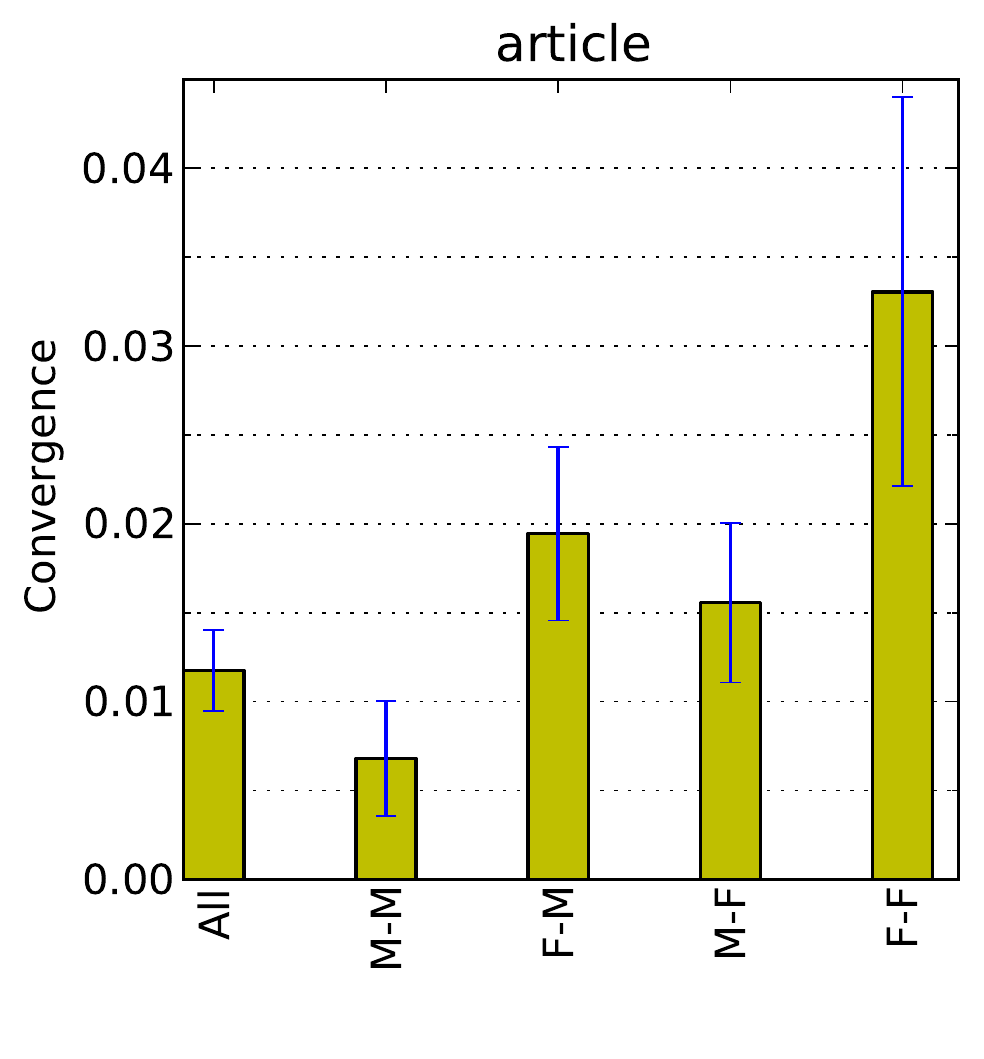}
\label{fig:gender2}
}
\caption{
Relation between \textit{\article} \accommodation  and imagined gender.
(a) compares cases when the \textbf{init}iator and \textbf{resp}ondent are \Male or \Female;  (b) compares types of gendered \textbf{initiator-respondent} relations: \Male-\Male, \Female-\Male, \Male-\Female, \Female-\Female . For comparison, the 
\textbf{All} 
bars represents the general \textit{\article} \accommodation (illustrated in Fig. \ref{fig:acc} as the difference between the two respective bars).
}
\end{figure*}

\subsection{\Accommodation and imagined relation}

\begin{figure*}[th!]
\centering
\subfigure[\textbf{F resp.}
minus \textbf{M resp.}]{
\includegraphics[width=\sumwidth]{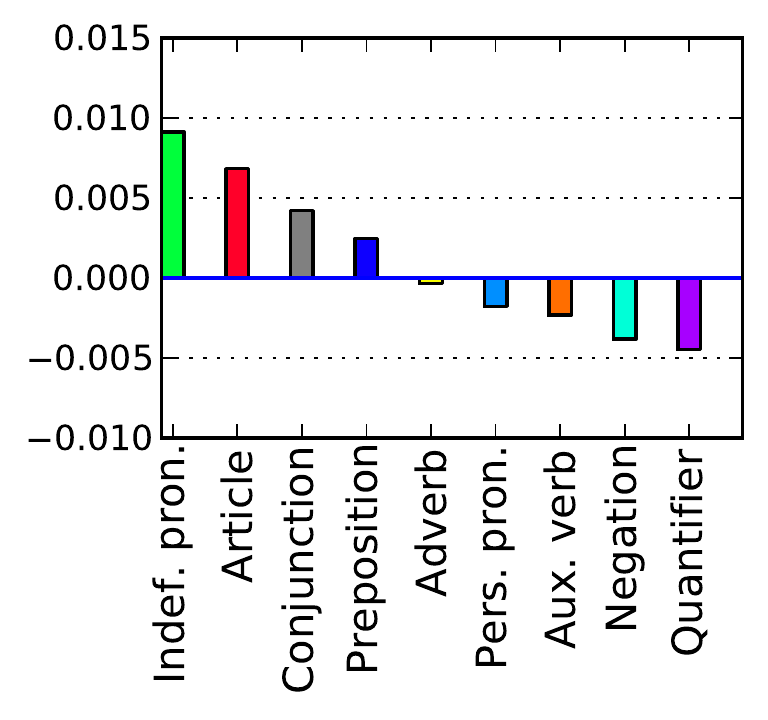}
}\subfigure[\textbf{F init.} 
minus
\textbf {M init.}]{
\includegraphics[width=\sumwidth]{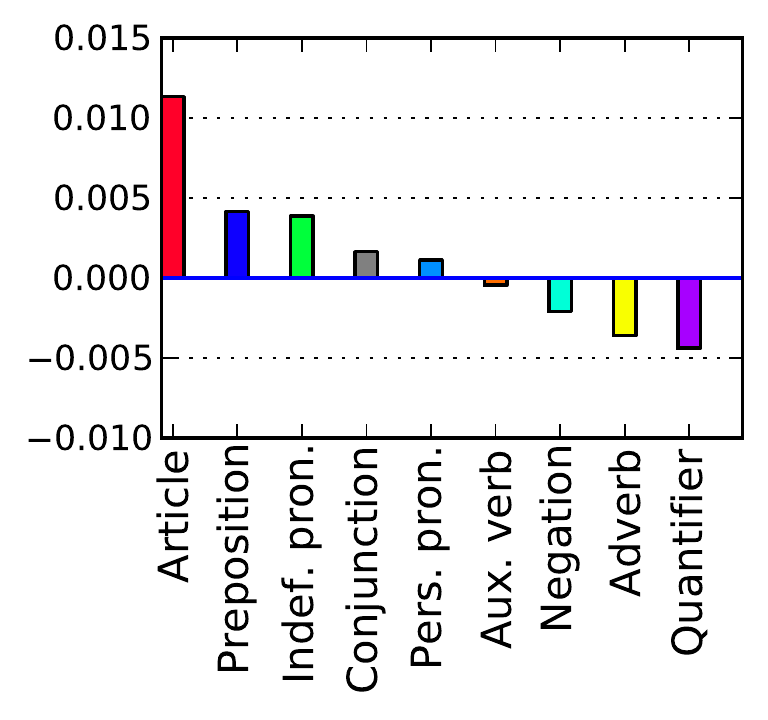}
} \sumfigsplitter
\subfigure[\textbf{\Lead}
minus
\textbf{\Second}]{
\includegraphics[width=\sumwidth]{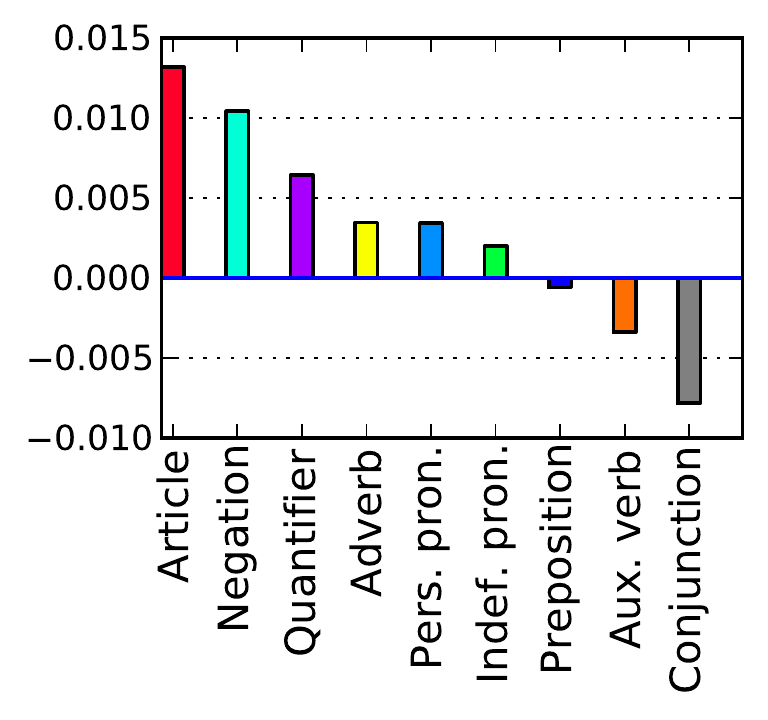}
}
\subfigure[\textbf{Quarrel} minus \textbf{No quarrel}]{
\includegraphics[width=\sumwidth]{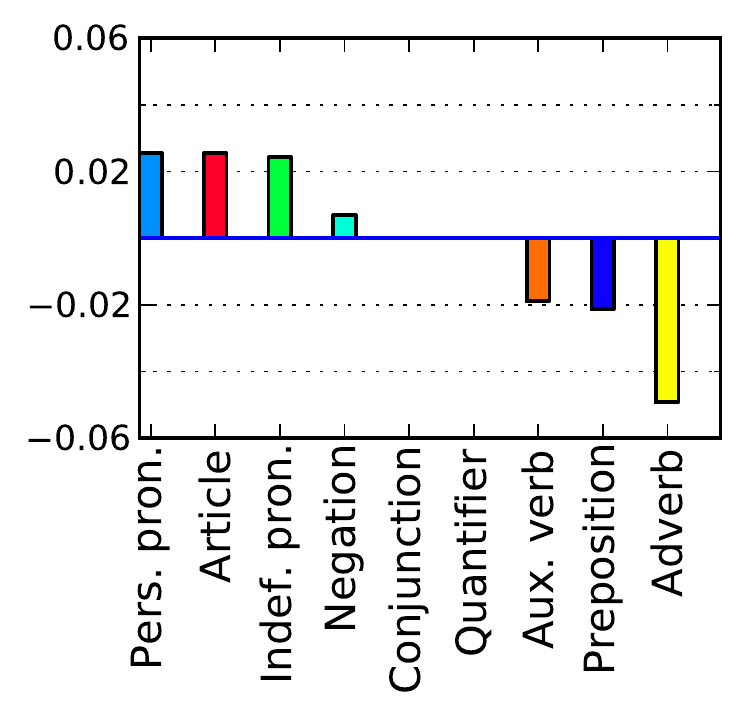}
}
\caption{\label{fig:diffs}
 Summary of the relation between  \accommodation
 and imagined gender (a and b), billing order (c), 
 and 
quarreling (d). 
 The bars represent the 
{\em difference} between the \accommodation observed
 in the respective cases; e.g., the 
\textbf{\article} 
(red)
bar in (a)
 represents the 
difference between the \textbf{F resp.} and the
 \textbf{M resp.} bars in Fig. \ref{fig:gender1}. 
In each plot, 
the trigger
 families are sorted according
to
 the respective difference,
but the color assigned to each family is consistent across plots.
The scale of (d) differs from the others.
}
\end{figure*}

We now
analyze how \accommodation patterns vary with the type of
relationship between the  (imagined) participants.
Note that, given the multimodal character of \accommodation,
treating each \dimension separately is the most appropriate way to
proceed,
since 
in past work, for
 the same experimental
factor (e.g., gender), different features converge differently (refer
back to \S\ref{sec:style}).
 For clarity of exposition, we  
discuss in detail only the  results for
 the \textit{\articles}
feature family;
but 
the results for
all \dimensions are summarized in Fig.  \ref{fig:diffs},
discussed later.

\paragraph{Imagined gender}

Fig. \ref{fig:gender1} shows how \accommodation on article
usage depends on the
gender of the initiator and 
respondent. 
Females are more influential than males: movie characters
of either gender
accommodate more to female characters  than to male characters
(compare the \textbf{F}emale \textbf{init}iator bar with the
\textbf{M}ale \textbf{init}iator bar,
statistically significant,
independent t-test,  $p<0.05$).
Also, 
female characters seem to accommodate slightly more to other
characters than male characters do (though not statistically 
significantly so
in our data).

We 
also compare
the amount of \accommodation between all the possible types of 
gendered initiator-respondent pairs involved
(Fig. \ref{fig:gender2}).  One 
can 
observe, for example, that male
characters 
adapt
less in 
same-gender situations (\Male-\Male conversations) than in 
mixed-gender situations 
(\Female initiator-\Male respondent),
while the opposite is true for
 female characters 
({\Female-\Female} vs. {\Male-\Female}).

Interpreting these results lies beyond the scope of this paper.  We note that these results could be a correlate of many
factors, such as the roles that male and female characters are typically
assigned in movie scripts.\footnote{A comparison to previously reported results on real-life gender effects 
is not straightforward, since they pertain to different features; \newcite{Ireland:2010} show that females match their linguistic style more than males, where style matching is averaged over the same 9 trigger families we employ (they do not report gender effect for each family separately).}

\paragraph{\Narrativeimportance
}

\begin{figure}[t]
\centering
\includegraphics[height=\fight]{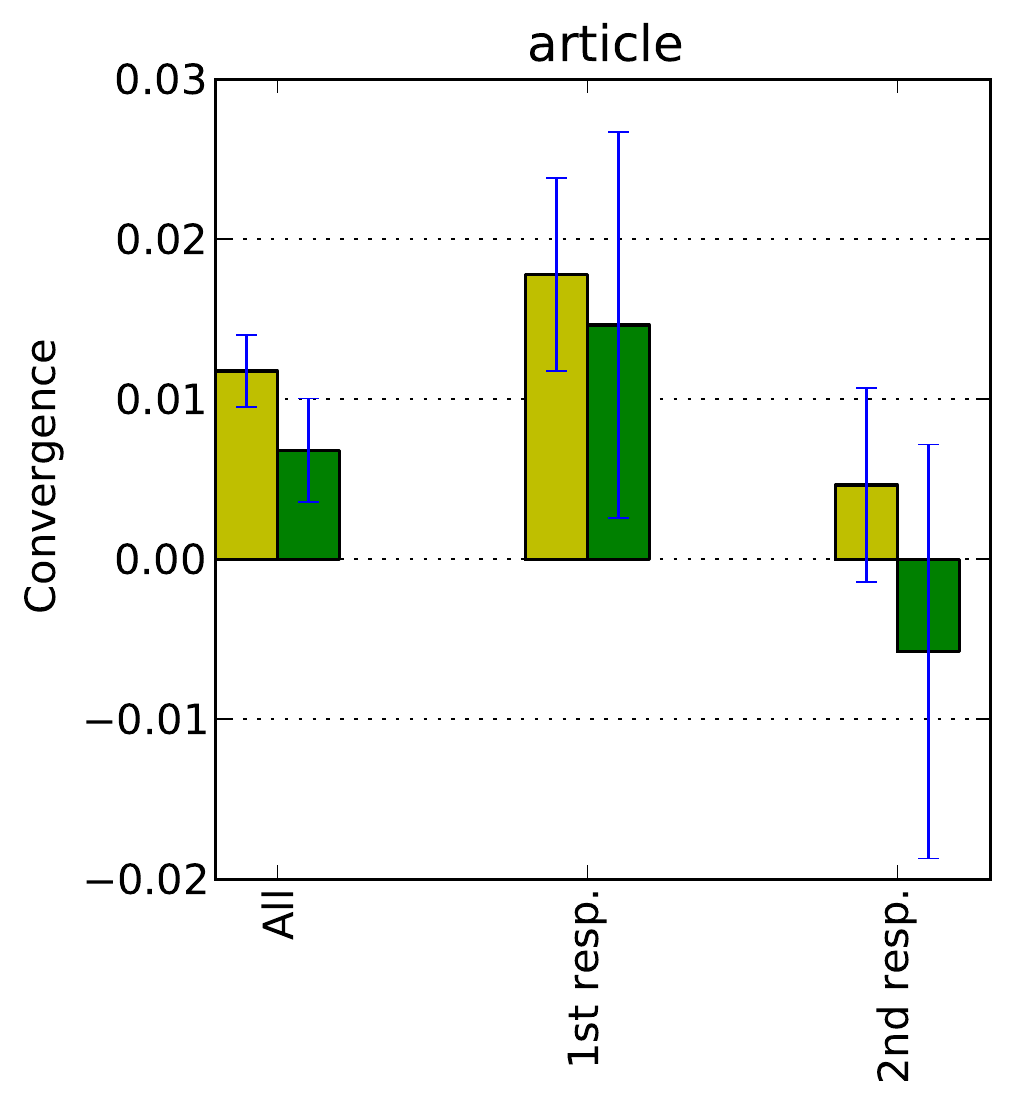}
\caption{
Comparison of
the
\accommodation of 
first-billed (lead)  characters to 
second-billed characters
(left bar in \textbf{\Lead} group) to that of 
second-billed characters to leads
(left bar in \textbf{\Second} group);
righthand bars
(dark green)
in each group show
results for Male-Male pairs only.
}
\label{fig:pos}
\end{figure}

Does the relative importance bestowed by the \scriptwriter to the
characters affect the amount of linguistic coordination 
he or she (nonconsciously) embeds in their dialogs?
Fig. \ref{fig:pos} shows that,
on average,
the lead character converges to
the
second-billed
 character 
more than  vice-versa 
(compare left bar in \textbf{\Lead} group with
left bar in \textbf{\Second} group).  

One possible confounding factor is that there is 
significant gender imbalance in the data 
(82\% of all lead characters are males, 
versus
only 51\% of the secondary characters).
Could
the observed difference
be
 a direct consequence of
the relation between gender and \accommodation discussed
 above?
The answer is no: the same qualitative observation holds if we restrict our
analysis to 
same-gender pairs (compare the righthand bars in each group in  Fig.
\ref{fig:pos}\footnote{
Figure \ref{fig:pos}
also shows that 
our \accommodation measure
does achieve negative values in practice, indicating divergence.
 Divergence is a rather common phenomenon which deserves attention in
 future work; see 
\newcite{Danescu-Niculescu-Mizil+al:11a} for an account.}). 

It would be interesting to see whether these results could be
brought to bear on previous results regarding the
relationship between social status and \accommodation, but such
interpretation lies beyond the scope of this paper, since the
connection between billing order and social status is not straightforward.

\begin{figure}[t]
\centering
\includegraphics[height=\fight]{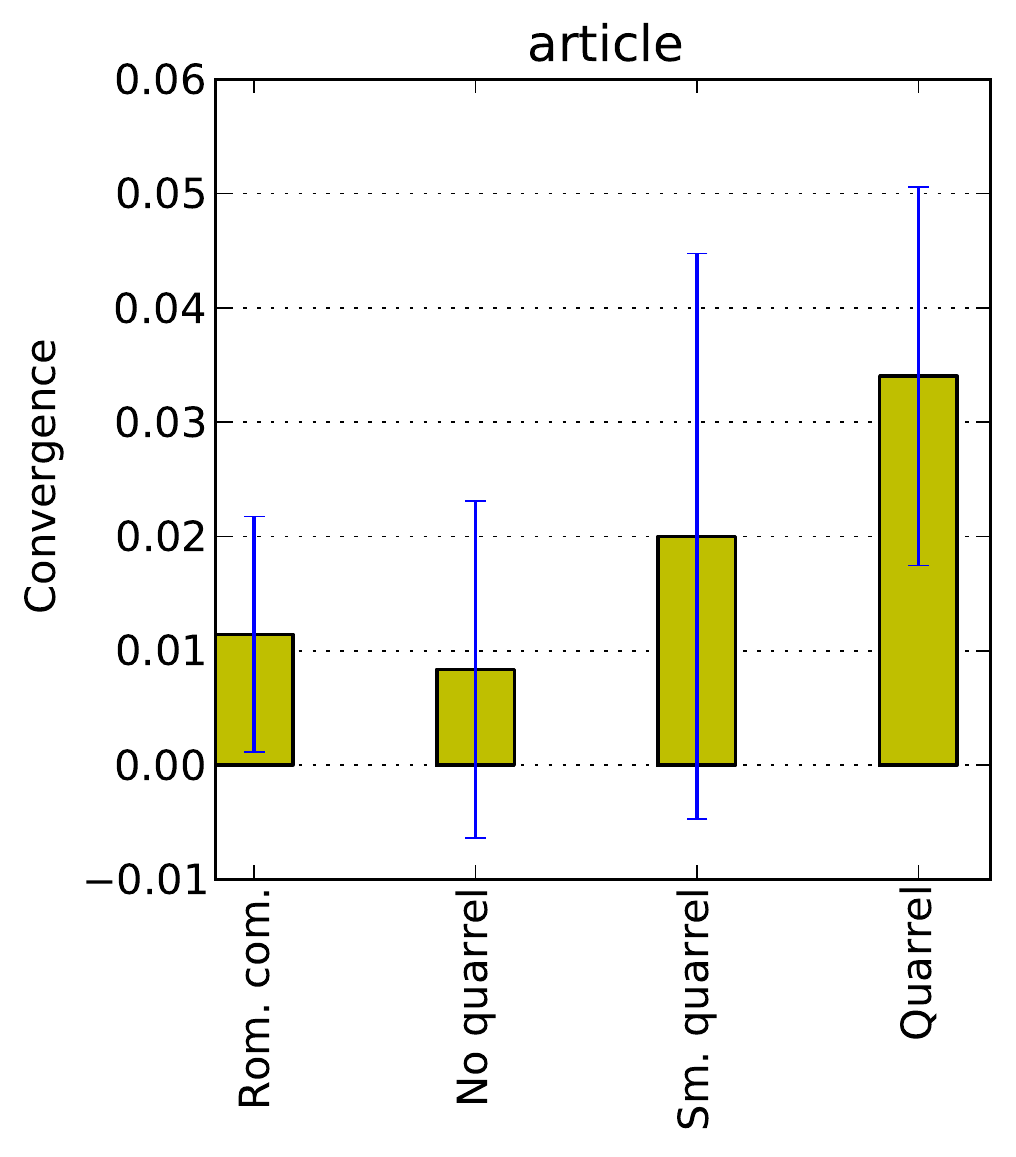}
\caption{Relation between contention and \accommodation.  
The third bar combines \textit{quarreling} and \textit{some
  quarreling} to ameliorate data sparsity.
For comparison, \textbf{Rom. com.}
shows \accommodation calculated on all the conversations of the 24
romantic-comedy pairs considered in this experiment.
}
\label{fig:quarreling}
\end{figure}

\paragraph{Quarreling}
The level of contention in conversations 
has also been
shown to be related to the amount of \accommodation
 \cite{Giles:2008,Niederhoffer:2002p2556,Taylor:2008p3340}.  To test
 whether this tendency
holds in the case of imagined 
conversations,
as a small pilot study, we 
manually classified the conversations between 24
main pairs of characters from romantic comedies\footnote{We chose the
romantic comedy genre since it is often characterized by some level of
 contention between the 
two people in the main couple.}
as:
\textit{quarreling}, \textit{some quarreling} and \textit{no
  quarreling}.  Although the experiment was too small in scale to
provide 
statistical
 significance, the results (Fig. \ref{fig:quarreling}) suggest that
 indeed the level of \accommodation is affected by the presence of
 controversy: 
\textit{quarreling} exhibited considerably more \accommodation for
articles than
the other categories  (the same holds for
personal and indefinite pronouns;
see
Fig. \ref{fig:diffs}).  Interestingly, the reverse is true for
adverbs;
there,  we observe divergence for contentious conversations and
\accommodation for non-contentious conversations 
 (detailed plot
omitted due to space
constraints).   This corresponds to 
Niederhoffer and Pennebaker's \shortcite{Niederhoffer:2002p2556}
observations 
made on 
real conversations in their study of the
Watergate transcripts: when the 
relationship
 between the two deteriorated,
Richard 
Nixon  converged more to John Dean on articles, but
diverged
on other 
features.\footnote{Adverbs were not included in their study.}

\paragraph{Results for the other features}
Our results
above
suggest some intriguing interplay between
\accommodation and 
gender, 
status, and level of hostility
in imagined dialogs, 
which may shed light on 
how
people 
(scriptwriters)
nonconsciously \textit{expect} \accommodation to
relate to 
such 
factors. 
(Interpreting
these 
sometimes
apparently counterintuitive 
findings is beyond the scope of this 
paper,
 but represents a fascinating direction for future work.)
 Fig. \ref{fig:diffs}
shows how the
nature of these relations depends on the \dimension considered.
The variation among families is
in
line
with the previous empirical results
on
the multimodality of
\accommodation in real conversations,
as 
discussed in \S\ref{sec:style}.

\section{Summary and future work}

We provide some insight into the causal mechanism behind \accommodation, 
a topic that has generated substantial scrutiny and debate for over 40
years \cite{Ireland01122010,Branigan:2010p3605}.  
Our work, along with 
\newcite{Elson+McKeown:2010}, advocates for the
value of fictional sources in the study of linguistic
and social
 phenomena. To stimulate such
 studies, we render our metadata-rich corpus of movie dialog public.

In \S\ref{sec:intro}, we 
described some 
practical
applications of a better understanding of the chameleon effect in language; it boils down to improving communication both between humans and between humans and computers. 
Also, our results on contention 
could be used to further automatic 
controversy detection \cite{Mishne+Glance:2006a,Gomez+Kaltenbrunner+Lopez:2008a}.
Moreover, if we succeeded in linking our results on
\narrativeimportance to relative social status, we might further the
development of  systems that can infer social relationships  in online social networks when conversational data is
present but other, more explicit cues are absent
\cite{Wyatt+al:2008a,Bramsen+al:2011a}.
Such systems could 
be valuable
to the rapidly expanding field of analyzing social networks.

\vspace{0.8cm}
{

\ifackfirstinitial
\newcommand{\firstname}[2]{#2.}
\else
\newcommand{\firstname}[2]{#1}
\fi

\paragraph{Acknowledgments} Many thanks for their help are due to
\firstname{Claire}{C} Cardie,   
\firstname{Catalin}{C} Draghici, 
\firstname{Susan}{S} Dumais,  
\firstname{Shimon}{S} Edelman, 
\firstname{Michael}{M} Gamon, 
\firstname{Jon}{J} Kleinberg, 
\firstname{Magdalena}{M} Naro\.zniak, 
\firstname{Alex}{A} Niculescu-Mizil, 
\firstname{Myle}{M} Ott, 
\firstname{Bo}{B} Pang, 
\firstname{Morgan}{M} Sonderegger, 
plus NLP seminar participants 
\firstname{Eric}{E} Baumer, 
\firstname{Bin}{B} Lu, 
\firstname{Chenhao}{C} Tan, 
\firstname{Lu}{L} Wang, 
\firstname{Bishan}{B} Yang,  
\firstname{Ainur}{A} Yessenalina, 
and
Marseille, 
and the anonymous reviewers
(who went far beyond the call of duty!).  Supported by NSF
IIS-0910664,  the
Yahoo! FREP program,
and a Yahoo! Key Scientific Challenges award.
}

\normalsize

\tiny

\end{document}